%% file: main.tex
\documentclass[10pt,twocolumn,letterpaper]{article}

\usepackage{iccv}
\usepackage{times}
\usepackage{epsfig}
\usepackage{graphicx}
\usepackage{amsmath}
\usepackage{amssymb}
\usepackage{enumitem}
\usepackage{multirow}
\usepackage{lipsum}
\usepackage{caption}
\usepackage{xspace}
\usepackage{paralist}

\input{macro.tex}

\usepackage[pagebackref=true,breaklinks=true,letterpaper=true,colorlinks,citecolor=blue,linkcolor=blue,bookmarks=false]{hyperref}

\usepackage[hang,flushmargin]{footmisc}
\newcommand{\mycaption}[2]{\caption{\textbf{#1.}~#2}}
\newcommand\blfootnote[1]{%
  \begingroup
  \renewcommand\thefootnote{}\footnote{#1}%
  \addtocounter{footnote}{-1}%
  \endgroup
}

\iccvfinalcopy % *** Uncomment this line for the final submission

\def\iccvPaperID{1682} % *** Enter the ICCV Paper ID here

\ificcvfinal\fi

\begin{document}

%%%%%%%%% TITLE
\title{Motion-Conditioned Diffusion Model for Controllable Video Synthesis \vspace{-2mm}}

\author{
Tsai-Shien Chen$^1$ \quad Chieh Hubert Lin$^1$ \quad Hung-Yu Tseng$^2$ \quad Tsung-Yi Lin$^3$ \quad Ming-Hsuan Yang$^{1,4}$\\
$^1$University of California, Merced \quad $^2$Meta \quad $^3$NVIDIA \quad $^4$Google Research\\
}

\twocolumn[{
\maketitle
\vspace{-1.em}
\renewcommand\twocolumn[1][]{#1}%
    \centering 
    \includegraphics[width=\linewidth]{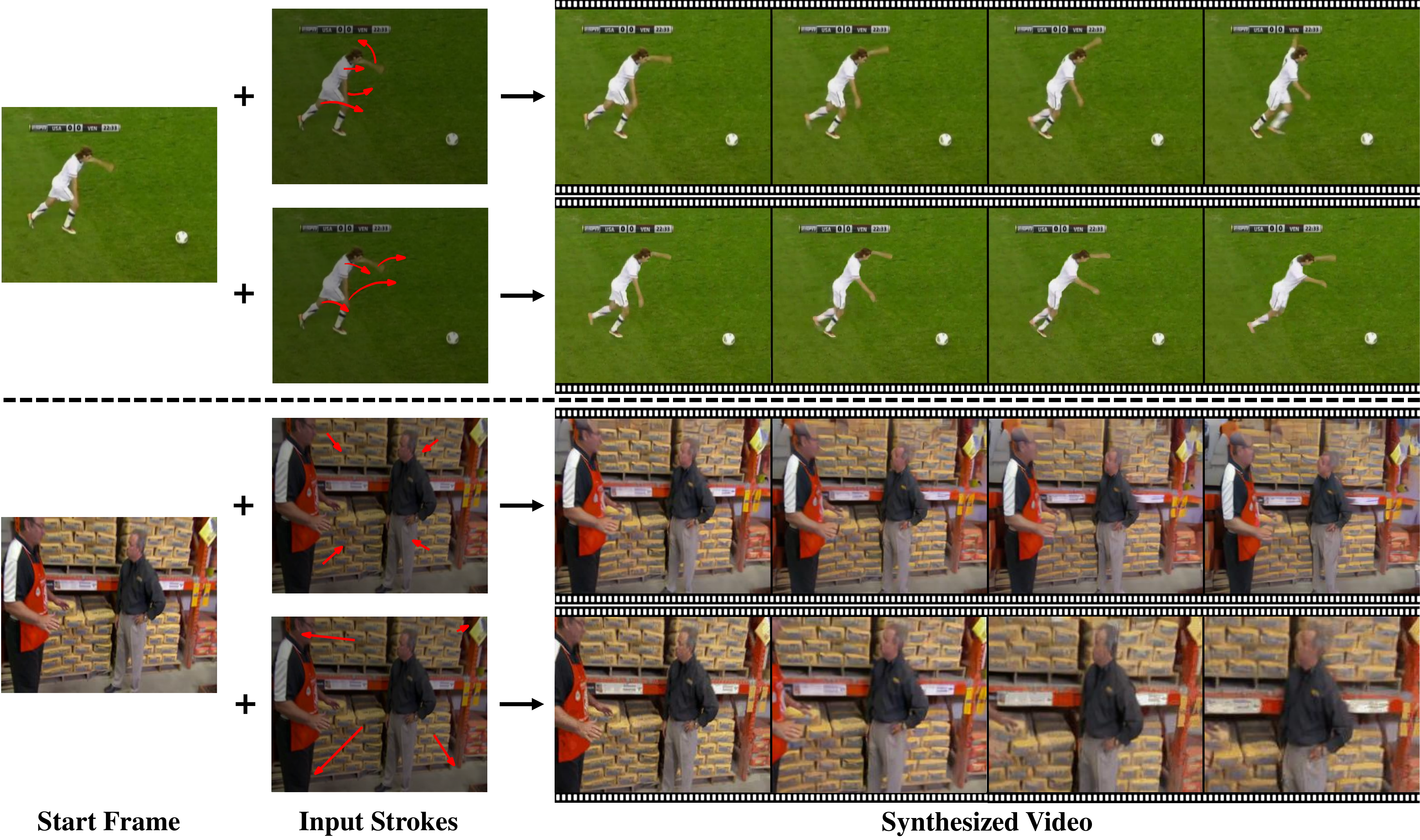}
    \vspace{-1.7em}
    \captionof{figure}{
    \textbf{MCDiff enables flexible and accurate motion control in high-quality video synthesis with diffusion models.}
    Given a start frame and a set of user-specified strokes ({\color{red} red} arrows), our proposed MCDiff synthesizes a video following the desired motion while preserving the content.
    % with user-specified contents and motions.
    %
    We show that MCDiff learns the concept of foreground and background flows, where the former specifies the motions of foreground objects (as top-two rows), while the latter controls the camera adjustments (\eg, zoom-in or zoom-out as bottom-two rows).
    % Through MCDiff, users can manipulate both the motion of the foreground objects (top two rows) and the camera viewpoint (bottom two rows,) such as zoom-in or zoom-out, by assigning background motion.
    } \label{fig:teaser}
    \vspace{+2mm}
}]

\blfootnote{Project page: \url{https://tsaishien-chen.github.io/MCDiff/}}

%%%%%%%%% ABSTRACT
\vspace{-.3em}
\begin{abstract}
    \vspace{-.5em}
    % Dang it, ChatGPT is too good
    %\hubert{
    Recent advancements in diffusion models have greatly improved the quality and diversity of synthesized content.
    To harness the expressive power of diffusion models, researchers have explored various controllable mechanisms that allow users to intuitively guide the content synthesis process.
    Although the latest efforts have primarily focused on video synthesis, there has been a lack of effective methods for controlling and describing desired content and motion.
    In response to this gap, we introduce MCDiff, a conditional diffusion model that generates a video from a starting image frame and a set of strokes, which allow users to specify the intended content and dynamics for synthesis.
    %
    %{\color{red}
    To tackle the ambiguity of sparse motion inputs and achieve better synthesis quality, MCDiff first utilizes a flow completion model to predict the dense video motion based on the semantic understanding of the video frame and the sparse motion control.
    Then, the diffusion model synthesizes high-quality future frames to form the output video.
    %\hubert{HY and I both think this is too long.}
    %
    %Our evaluation of MCDiff on multiple large-scale video datasets demonstrates that the model is able to reflect the specified strokes while maintaining high visual quality.
    We qualitatively and quantitatively show that MCDiff achieves the state-the-of-art visual quality in stroke-guided controllable video synthesis.
    Additional experiments on MPII Human Pose further exhibit the capability of our model on diverse content and motion synthesis.
\end{abstract}

%%%%%%%%% BODY TEXT
\input{1_introduction.tex}

\input{2_related_work.tex}

\input{3_methodology.tex}
\input{4_experiment.tex}
\input{5_conclusion.tex}

{\small
\bibliographystyle{ieee_fullname}
\bibliography{egbib}
}

%\clearpage
%\onecolumn
%\appendix
%\input{6_appendix}

\clearpage
\onecolumn
\appendix

\begin{center}
\section*{Appendix}
\end{center}

%%%%%%%%% BODY TEXT
\input{6_appendix}

\end{document}

%% file: macro.tex
\usepackage{color}
\definecolor{crimson}{rgb}{0.86, 0.08, 0.24}
\definecolor{gray}{rgb}{0.5,0.5,0.5}
\definecolor{green}{rgb}{0, 1, 0}
\definecolor{orange}{rgb}{1, 0.5, 0}
\definecolor{mahogany}{rgb}{0.75, 0.25, 0.0}
\definecolor{purple}{rgb}{0.6, 0, 0.6}
\definecolor{darkgreen}{rgb}{0, 0.4, 0}
\definecolor{frenchblue}{rgb}{0.0, 0.45, 0.73}
\definecolor{red}{rgb}{1,0,0}
\definecolor{yellow}{rgb}{0.9,0.9,0}
\definecolor{magenta}{rgb}{1,0,1}
\definecolor{pink}{rgb}{1,0.412,0.706}
\definecolor{newgreen}{rgb}{0, 0.6, 0.2}
\definecolor{cyan}{rgb}{0,1,1}

\long\def\ignorethis#1{}
\newcommand {\hubert}[1]{{\color{mahogany}\textbf{Hubert: }#1}\normalfont}
\newcommand {\ts}[1]{{\color{red}\textbf{TS: }#1}\normalfont}

\newcommand{\comment}[1]{}

%% file: 1_introduction.tex
\vspace{-6mm}
\section{Introduction}

%\hubert{
Controllable video synthesis that allows users to specify desired contents and motions has numerous applications in the areas such as visual effects synthesis, video editing, and animation.
% 
% Specify what is challenging
%Such a problem remains challenging for modern generative models due to the diversity of contents and ill-conditioned control signals (\eg, the sparse strokes we studied in this %paper).
% TS: can our model solve the diverse content problem?
% 
Although the emerging diffusion models~\cite{ddpm,ldm} have made significant strides in video synthesis, the degree of controllability remains limited to coarse-grained classes~\cite{VDM} or text prompts~\cite{text2live,scenescape,imagenvideo,makeavideo}.
% While the recently emerging diffusion models~\cite{ddpm,ldm} have shown remarkable breakthroughs in video synthesis, its controllability remains coarse-grained classes~\cite{VDM} or text prompts~\cite{text2live,scenescape,imagenvideo,makeavideo}.
% 
%\hy{finer-grained over? motion?} \hubert{Is it better now with specifying the use case?}
% hubert{I don't quite get it, you mean we better specify the control is on the motion?} \hy{fine-grain can be used to describe different aspects (class, motion, ...), I assume we are emphasizing motion right?} \ts{I believe it should be finer-grained control (both start frame and strokes can provide more accurate control on video synthesis compared to texts or class labels)}
%\hubert{
The observation piques our interest in more finer-grained user control over the video synthesis procedure, specifically, how to enable users precisely specify the anticipated subjects and motions within the synthetic video.
%}
% 
% \hubert{In particular, we study controlling the video dynamics with strokes.}
%}
%Conditionally synthesizing a video to animate an object with a specific motion is a challenging yet practical task.
%
%It has a wide range of applications for video creators, including visual effects editing, animation production, etc.
%
%Recently, denoising diffusion models (DDM)~\cite{ddpm,ldm} have been shown to achieve remarkable breakthroughs in video synthesis conditioned on class attributes~\cite{VDM} and text prompts~\cite{imagenvideo,makeavideo,text2live,scenescape}.
%
%In this paper, we aim at exploring a different \hubert{``different'' is a vague word} level of controllability for video synthesis.
%
%\hubert{
%\hy{Split the following to a new paragraph?} \ts{Will the paragraphs be too short after splitted? If not, I think it is better to split them.}

Considering that a video can be seen as a combination of content and motion components~\cite{MoCoGANHD,MoCoGAN}, as in Figure~\ref{fig:teaser}, we design an interface that enables users to specify the content using a reference image as the start frame of the video and control the motion with a set of stroke inputs.
% Considering that a video can be expressed as a combination of content and motion components~\cite{MoCoGANHD,MoCoGAN}, as illustrated in Figure~\ref{fig:teaser}, we design an interface where the content is specified with a reference image as the start frame of the video, and the motion is controlled by a set of stroke inputs.
%
% As two examples in Figure~\ref{fig:teaser}, t % Double referring
These strokes can depict the desired motions of both the foreground subjects and the camera adjustments (\eg, zooming or shifting viewpoint) throughout the video.
Our goal is to synthesize a video sequence that faithfully reflects the specified content and motion information.
While previous works~\cite{ipoke,II2V} show exploratory results on a similar problem with simple datasets, such as TaiChi-HD~\cite{taichi} and Human3.6M~\cite{human36m}, these datasets remain simple (single subject with monotonous actions in a dull background), and the synthesis quality remains preliminary as shown in Section~\ref{sec:sota}.
As an attempt to address the visual quality issues by leveraging the expressiveness of diffusion models, we implement a single-stage conditional diffusion model that directly synthesizes a video based on the start image frame and the stroke inputs.
However, as shown in Section~\ref{sec:single-stage}, this naive approach leads to dissatisfactory results.
We hypothesize the failure is caused by the challenging nature of the task in two aspects: (a) the ambiguity exhibited by the sparse strokes, and (b) the requirement of semantic understanding for realistic video synthesis.
For instance, consider a case where an input stroke indicates moving the hand of a character upward.
The stroke is only attached to a single pixel, but the action should propagate through all pixels corresponding to the hand. 
Meanwhile, the arm of the character should have a coherent motion that is unnecessarily in the same direction and magnitude as the stroke.
Such a problem becomes drastically more difficult as the diversity of objects, scenes, and activities within the dataset increases.
Thus, it denies the diffusion model from converging and producing reasonable videos at high quality.

To reduce the ambiguity and difficulty exhibited in the single-stage end-to-end learning task, we present \emph{\underline{M}otion-\underline{C}onditioned \underline{Diff}usion Model}, abbreviated as MCDiff, a two-stage framework that breaks down the task into two sub-problems: sparse-to-dense flow completion and future frame prediction.
The flow completion task aims to convert the input strokes (equivalent to sparse flows) into denser flows that represent the motion within a video accordingly based on the semantic meaning of the input video frame.
The following future-frame prediction model generates the next video frame through a conditional diffusion process, based on the current frame and the predicted dense flows.
Lastly, the two networks are end-to-end fine-tuned into a coherent and synergetic model.
We observe the two-stage model stabilizes the training and successfully leverages the excellence of diffusion models, achieving eminent visual quality while faithfully following the instructions made with the input strokes.

We evaluate MCDiff on three large-scale video datasets.
The experiments on the standard benchmarks~\cite{human36m,taichi} quantitatively and qualitatively show that MCDiff achieves state-of-the-art visual quality over previous stroke-guided controllable video synthesis methods.
We then further examine the limit of our method on MPII Human Pose~\cite{mpii} dataset, which is a collection of videos with over 400 human activities captured in various conditions and camera settings.
%}
% Noticeably, as illustrated in Figure 2, MoCoDiffusion shows the first attempt at controllable video synthesis targeting to predict the video sequences in MPII Human Pose~\cite{mpii} dataset which collects videos with more than 400 dramatic human body motion, such as mountain biking and skiboarding, and captured by various camera movement techniques.
%\hubert{
The experiments demonstrate the capability of the proposed model on synthesizing videos with diverse contexts and motions.
%
%\hubert{
Moreover, MCDiff discovers the strokes on the foreground subjects indicate object movements, while the strokes on the background scene represent camera adjustments.
Such behavior enables more flexible control and realistic camera trajectories for video synthesis and editing.

%% file: 2_related_work.tex
\begin{figure*}[t]
    \centering
    \includegraphics[width=\linewidth]{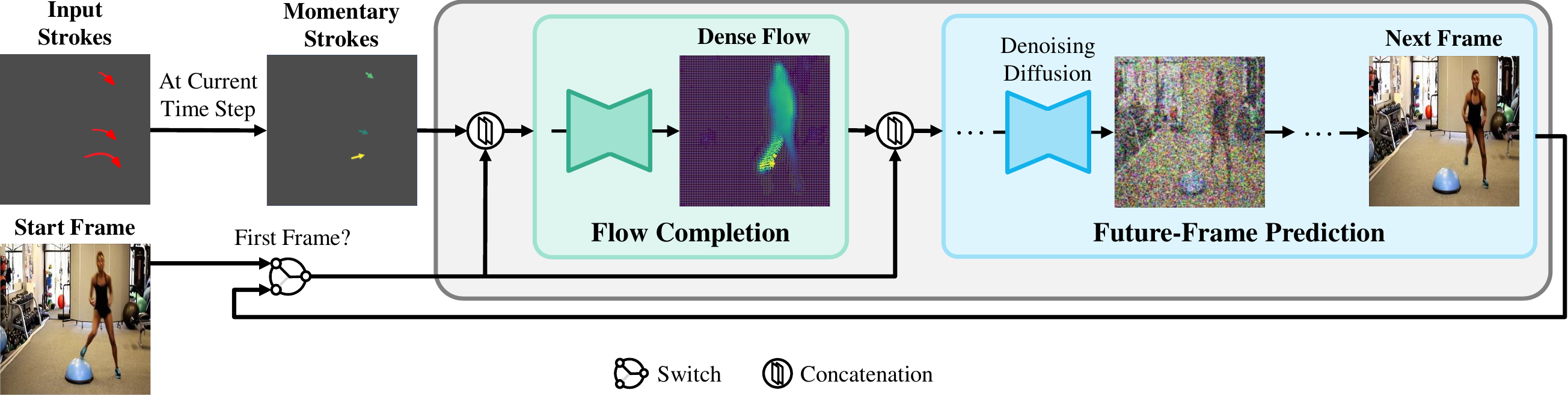}
    %\vspace{-7mm}
    \mycaption{Overview}{
        %\hubert{
        MCDiff is an autoregressive video synthesis model. 
        For each time step, the model is guided by the previous frame (\ie start or previously predicted frame) and the momentary segment of input strokes (marked as colored arrows, a brighter color indicates a larger motion).
        Our flow completion model first predicts dense flows representing per-pixel momentary motion. 
        Then, the future-frame prediction model synthesizes the next frame based on the previous frame and the predicted dense flow through a conditional diffusion process.
        Finally, the collection of all predicted frames forms a video sequence adhering to the context provided by the start frame and the motion specified by the strokes.
        %}
        % MCDiff synthesizes a video in an autoregressive fashion.
        % %
        % To generate a video frame, the flow completion model first predicts dense motion from the input strokes.
        % %with the guidance of the reference image.
        % %
        % We then synthesize a future frame based on the current frame and the predicted dense flow through a conditional diffusion model.
        % %
        % The arrows represent the point movements, colored with their normalized motion magnitudes (\ie {\color{purple} purple} for the slightest motion while {\color{yellow}yellow} for the largest one.)
    }
    \label{fig:architecture}
    %\vspace{-3mm}
\end{figure*}

\section{Related Works}

%\subsection{Controllable Diffusion Models For Content Synthesis}
%\hubert{
\noindent\textbf{Controllable Diffusion Models for Content Synthesis.}
Many recent works propose novel controlling mechanisms tailored for the image or 3D-shape diffusion models. 
These methods aim to provide users control over the process of synthesizing novel contents, typically the appearances or shapes of the synthesis target. 
For image diffusion models, these control mechanisms include texts~\cite{ediffi,dalle2,ldm,imagen}, image references~\cite{customdiffusion,dreambooth}, sketches of segmentation maps~\cite{sdedit,ldm}, silhouettes~\cite{voynov2022sketch}, or scene graph~\cite{yang2022diffusionsg}.
Similar controlling mechanisms can also be found in 3D-shape diffusion models, \eg, texts~\cite{magic3D,dreamfusion}, image references~\cite{realfusion}, and segmentation maps~\cite{pix2pix3D}.
In this paper, we present another controlling mechanism for diffusion models achieving finer-grained control for video synthesis.
%}
%\hubert{
%Besides the image-based diffusion model, 

%In particular, several recent works apply diffusion models for 3D motion modeling.
%
%Tevet~\etal~\cite{MDM} models human motion, but the model is only coarsely controllable with text instructions.
%
%Raab~\etal~\cite{SinMDM} proposes to generate diverse motions from a single motion demonstration, but the method requires carefully designed and calibrated motions for a specific 3D object.
%}

\noindent\textbf{Video Diffusion Models.}
Recently, several approaches based on diffusion models have been developed for synthesizing temporally consistent videos~\cite{harvey2022flexible,imagenvideo,VDM,makeavideo,yang2022diffusion}.
The majority of these exploratory works either focus on unconditional video synthesis~\cite{VDM} or coarse-grained control~\cite{imagenvideo,makeavideo} over the synthesis process with text prompts.
A few most recent research explores finer-grained control over the diffusion process with different approaches, such as editing a video by analogizing it to different art styles~\cite{esser2023structure}, modifying the motion with text prompts~\cite{dreamix}, changing the main subject in the video~\cite{tuneavideo}, or extrapolating the motion from the demonstration~\cite{sinfusion}.
However, these approaches require an input video to provide the initial motion, which makes the controlling mechanisms unable to provide finer adjustments for the desired target motions.

A number of works treat video synthesis as the future-frame prediction task, which aims to synthesize videos based on an input image. 
Such a problem requires synthesizing appropriate motions based on the content clue from the image.
The prior methods~\cite{hoppe2022diffusion,kim2022diffusion} typically do not take extra input conditions and hence lack mechanisms to guide the motions of the target video.
Therefore, our controlling mechanism can be seen as an extension of the future-frame prediction task, as the workflow of many real-world designers and editors requires accurate and precise control to achieve their desired motions in the final product.
% 
%Talking head video synthesis~\cite{bigioi2023speech,facevid2vid} is another sub-field in video synthesis with one or many image reference(s).
% 
%Such a task uses human voice audios to guide the motion synthesis process, however, this is not applicable to the general motions in most human activities we tackled in this work.

\noindent\textbf{Controllable Video Synthesis.}
Numerous video synthesis methods exploit different types of controlling mechanisms.
Motion transfer~\cite{taichi,monkeynet,siarohin2021motion} or image reenactment~\cite{face2face} is a class of methods that extract motion from a reference video and use it to drive a still image.
These methods require video demonstration, thus making it more difficult to achieve fine-grained and accurate control. 
Playable video synthesis ~\cite{playableenv,playablevideo} learns a set of input instructions, such that the users can interactively control the spatial location of certain subjects in the scene and produce temporally coherent videos. 
However, these learned instructions remain simple coarse-grained actions and only control a single subject shared among all training data.

A few prior methods~\cite{clicktomove,ipoke,II2V,controllable} also adopt stroke inputs as their control mechanism, which are closely related to our study.
Hao~\etal~\cite{controllable} and C2M~\cite{clicktomove} both explicitly learn to model video motions by predicting dense optical flow maps and warp features following the flow maps for future-frame prediction.
However, such a warping operation often creates unnatural distortions that harm the visual quality.
% and requires accurate motion prediction which is impractical for videos with large motion.
% 
% In contrast, \hubert{contrast what!?}
% the most recent work on stroke-based interactive image-to-video synthesis, % repeating
On the other hand, II2V~\cite{II2V} and iPOKE~\cite{ipoke} compress videos into dense latent space and learn to manipulate these latent variables with a recurrent neural network~\cite{GRU}.
% learn an implicit motion representation to manipulate the image latent variables and synthesize the video through a recurrent neural network~\cite{GRU}.
%
%In Section~\ref{sec:sota}, we practically show this type of framework usually fails to handle the input strokes with large motions.
%
% Nonetheless, without leveraging the powerful diffusion model~\cite{ddpm,ldm} for content synthesis, the existing methods fail to synthesize high-quality videos.
%\hubert{
However, as shown in Figure~\ref{fig:ipoke}, the quality of these methods remains preliminary, and the generated outcomes often contain unnatural distortions, resulting in subpar visual quality.
%}
%To better achieve video synthesis with complex and large motion, our MCDiff explicitly models the video motion by predicting a dense flow map.
%\hubert{(a) These two sentences have no connections. (b) To claim such a motivation is true, you will need to do an ablation, I would suggest just referring to the empirical results.}
%
%But unlike Hao~\etal~\cite{controllable} and C2M~\cite{clicktomove}, we train our future-frame predictor in a data-driven manner to lessen the dependency on accurate motion prediction.
%\hubert{What do you mean by this? Aren't they data-driven? And why are you discussing this two times (see above)?}
%
% Moreover, we leverage the powerful image synthesis capability of the diffusion model to achieve better video frame synthesis.
% \hubert{This is unnecessary.}

%\hubert{
%A few prior methods~\cite{clicktomove,ipoke,II2V,controllable} also adopt stroke inputs as their control mechanism. 
%Hao \etal~\cite{controllable} proposes a framework that explicitly warps features following the strokes, such warping operation often creates unnatural distortions that harm the visual quality.
% 
%C2M~\cite{clicktomove} models the coarse-grained object-level displacements within a scene using a relation graph and GSN.
% 
%II2V~\cite{II2V} and iPoke~\cite{ipoke} use autoencoders to extract dense latent variables, then manipulate such latent variables with a motion model conditioned on stroke inputs.
% 
%In Section{???}, we show this type of method often ????
%}

%% file: 3_methodology.tex
\section{Method}
\label{sec:method}

%\hubert{\paragraph{Overview.}
MCDiff takes two types of inputs, a start frame image $x_0$ representing the content of the video, and a set of strokes controlling the motion. 
These strokes are interpreted as a set of sparse flows $\mathcal{S} = \{s_{1 \to 2}, ..., s_{(n-1) \to n} \}$, where each of the sparse flows $s_{i \to (i+1)}$ are the momentary displacements of the controlled pixels from the time step $i$ to $i+1$.
% which are 2D motion fields indicating the momentary displacement of pixels from time step $i$ to $i+1$.
%
Based on these inputs, our goal is to synthesize an $n$-frame video $\mathcal{X} = \{x_1, ..., x_n\}$ whose content and motion follow the input conditions.
%}
% Given a start frame $x_0$ and a set of strokes $\mathcal{S}$, our goal is to generate an $n$-frame video $\mathcal{X} = \{x_1, ..., x_n\}$ whose context and motion match the input conditions. 
% the following $n$ frames $\mathcal{X} = \{x_1, ..., x_n\}$ to compose of a video sequence whose context and motion match the input conditions. % This sentence is kind of stating the n-frame and video are not quite equivalent
%
% which can be decomposed in time-domain as $\{\mathcal{S}_{0 \to 1}, ..., \mathcal{S}_{n-1 \to n}\}$

%\hubert{
%\hubert{
In practice, we find that directly driving a conditional diffusion model with the sparse flows as inputs results in suboptimal performance, as shown in Section~\ref{sec:single-stage}.
%
%\ts{
Such an observation inspires the development of a two-stage framework with an additional module to tackle the ambiguity and difficulty of the sparse flow inputs.
%}
% In practice, we observe directly driving the conditional diffusion model with sparse flows yields dissatisfactory results, as shown in Section~\ref{sec:direct-sparse}.
% 
% To address the issue, we design a two-stage framework to facilitate learning.}
% 
Specifically, as illustrated in Figure~\ref{fig:architecture}, we synthesize a video in an autoregressive manner and accomplish the video frame synthesis by two components: flow completion $F$, and future-frame prediction $G$.
To generate a video frame $x_{(i+1)}$, we first utilize a flow completion model to predict a dense flow map $d_{i \to (i+1)}$ based on the current video frame $x_i$ and the sparse flow $s_{i \to (i+1)}$.
%Finally forming a set of dense flow maps $\mathcal{D} = \{d_{1 \to 2}, ..., d_{(n-1) \to n} \}$.
% 
Subsequently, we design a conditional diffusion model to generate the future frame $x_{(i+1)}$ based on the current frame $x_i$ and the predicted dense flow $d_{i \to (i+1)}$.
%}
%
In Section~\ref{sec:motion_distill}, we first introduce the annotations of video dynamics and then dive deeper into the design of the flow completion model in Section~\ref{sec:flow_completion}, and the future-frame prediction module in Section~\ref{sec:future_frame}.
%}
Lastly, we end-to-end fine-tune the two networks into a coherent and synergetic model in Section~\ref{sec:end2end}.

\comment{
To this end, we first decompose the video synthesis into future-frame prediction tasks.
To synthesize a future frame $x_{i+1} \in \mathcal{X}$, we learn a flow completion model (Section~\ref{sec:flow_completion}) to explicitly model a video motion by predicting a dense flow map $\mathcal{D}_{i \to i+1}$ based on the input sparse motion controls $\mathcal{S}_{i \to i+1}$ (decomposed $\mathcal{S}$ on time domain).
The flow completion model is trained on pairs of the videos and the extracted video dynamics (Section~\ref{sec:motion_distill}.)
Finally, we introduce a conditional diffusion model as the future-frame predictor (Section~\ref{sec:future_frame}) to generate the future frame $x_{i+1}$ conditioned on the current frame $x_i$ and the predicted dense flow map $\mathcal{D}_{i \to i+1}$.
The overall framework architecture is plotted in Figure~\ref{fig:architecture}. % this is not an architecture
}

\begin{figure}[t]
    \centering
    \includegraphics[width=\linewidth]{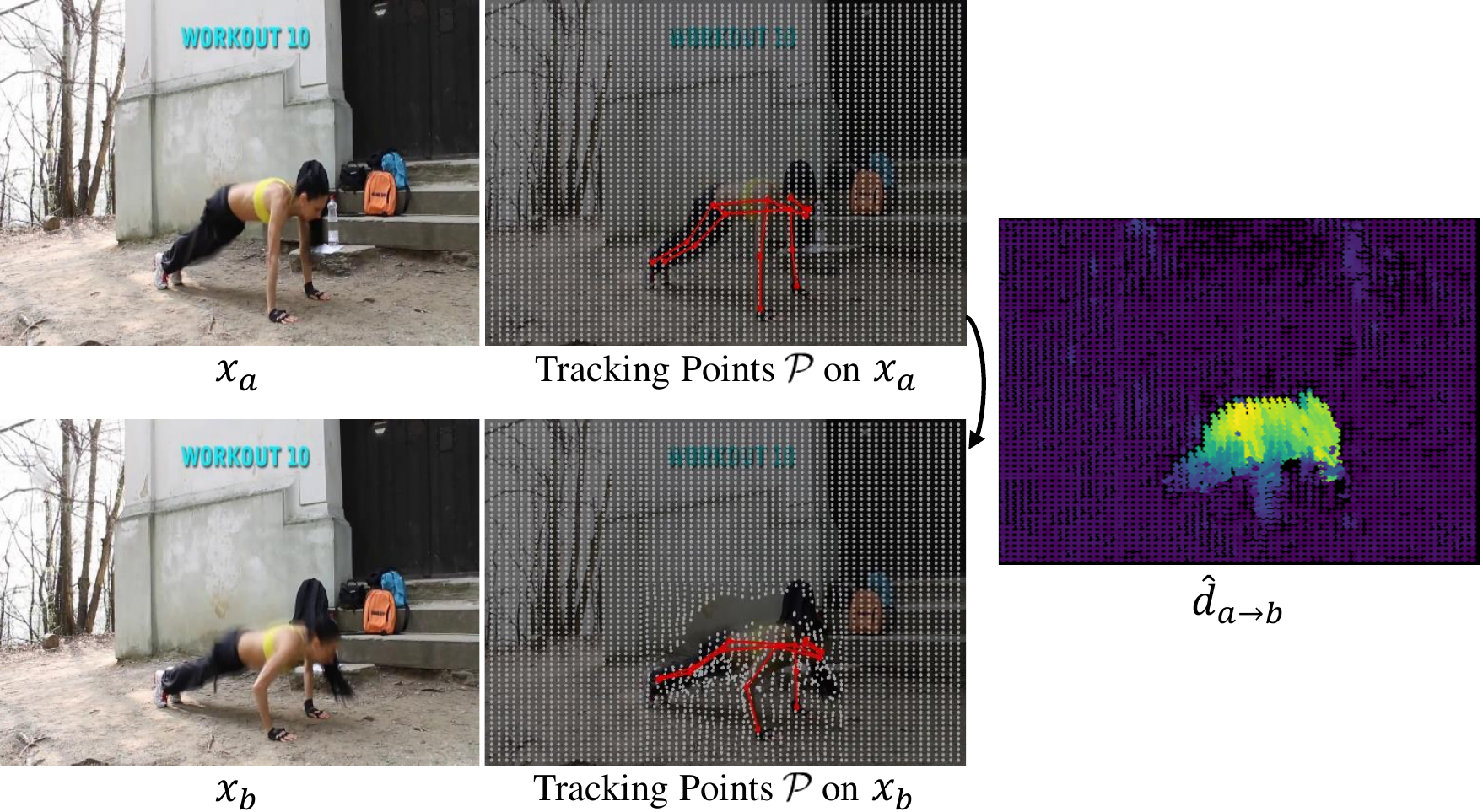}
    %\vspace{-7mm}
    \mycaption{Annotations of Video Dynamics}{
        %\ts{
        We express video dynamics by tracking both the keypoints ({\color{red} red}, marked with body skeletons for better visualization) and a grid of general points ({\color{gray}gray}).
        With the trajectories of the tracking points throughout a video, we can easily yield the dense flow map $\widehat{d}_{a \to b}$ between two arbitrary frames $(x_a, x_b)$.
        %}
        %
        %Our motion description can express informative long-term dynamics for complex foreground motions and viewpoint shifting. 
        %\hubert{I haven't revised this. The caption needs to focus on what to see in the figure, and why is it important. The current one is too descriptive. Like, this is a rock, I know it is a rock, but why is it special and worth my time watching it?}
    }
    \label{fig:motion_distill}
    %\vspace{-3mm}
\end{figure}

\subsection{Annotations of Video Dynamics}
\label{sec:motion_distill}
%\hubert{
Flow is an intuitive representation for expressing video dynamics.
A dense flow marks pixel-wise motion direction from one video frame to another, describing the incurred motion of the corresponding pixel during the time elapsed. 

As in Figure~\ref{fig:motion_distill}, to obtain the dense flows of a video, we scatter an array of tracking points $\mathcal{P}$ on the image frame, then retrieve the trajectories of these tracking points throughout a video by both general points and keypoints tracking algorithms~\cite{particle,hrnet}.
We aggregate the two results by overriding the trajectories of the general points with the trajectories of the keypoints, as the latter tends to be more accurate with the human shape prior.
%, while the general points tracking algorithm is a all-purpose model for any scene.
Eventually, with the densely annotated trajectories throughout the video, we can easily yield a dense flow map between two arbitrary frames $(x_a, x_b)$, denoted as $\widehat{d}_{a \to b}$.

\subsection{Flow Completion Model}
\label{sec:flow_completion}
Given the current frame $x_i$ and the momentary sparse flows $s_{i \to (i+1)}$, the flow completion model $F$ aims at predicting the dense flow map $d_{i \to (i+1)}$.
%
% The flow: sparse into 2D => why is this operation has a bit of issue => how we address the issue
%\hubert{
We first reformat the sparse flows $s_{i \to (i+1)}$ into a 2D map where each pixel indicates the user-specified motion from the time step $i$ to $(i+1)$. This operation introduces plenty of missing values due to the sparsity of the flows $s_{i \to (i+1)}$. We fill in the missing values with a shared and learnable embedding~\cite{bert,MAE} to indicate the absence of a user-specified motion in these pixels.
The 2D map is then concatenated with $x_i$ forming the input data of $F$, implemented as a UNet.
% Our flow completion model is a UNet that takes the concatenated current frame and 2D flow map as its input.
%}
% We first express the sparse flow $s_{i \to i+1}$ as a 2D motion field, where the vectors at the pixels with the controlled signals are the user-specified displacements while the rest of locations are filled with a mask token~\cite{bert,MAE}, a shared and learnable vector to indicating the absence of the control signal.
% 
%\hubert{what are "controlled" pixels and mask tokens? you never define it.}
%
% % Hubert: two? which two?
% We then concatenate two inputs and generate the dense flow map through a UNet.

We train $F$ in a self-supervised manner.
That is, for each pair of frames $(x_a, x_b)$ as training data, we simulate the user-specified sparse flows $s_{a \to b}$ by downsampling the extracted dense flow $\hat d_{a \to b}$.
To better preserve meaningful video dynamics within a sparse set of flows, we prioritize the sampling of the flows corresponding to the keypoint pixels, as they are more critical to representing the subject movements within a video.
Moreover, we sample the flows of the keypoints using the flow magnitude as the sampling probability, considering that a flow with a larger magnitude is usually more representative.
On the other hand, we also randomly sample the flows of general tracking points to represent the motion of other objects and background movement due to camera adjustments.

%}
\comment{
During the training on a pair of frames $(x_a, x_b)$, we automatically generate the sparse flows $s_{a \to b}$ by undersampling the extracted dense flow $\hat d_{a \to b}$.
To better simulate human behavior, which usually specifies the motions on the representative object landmarks with larger motions, we sample a portion of the landmarks.
We also sample a fixed number of motions from a grid of sampled points to imitate the users' interest in the control of background or camera viewpoint movement.
%
%sample a pair of frames $x_a$ and $x_b$ where $a < b$ and a set of sparse flows $d_{a \to b}$ sampled from 
}

With the simulated sparse flows $s_{a \to b}$ as inputs, we then supervise $F$ with an MSE loss between $\hat d_{a \to b}$ and the predicted dense flows $d_{a \to b}$.
However, the distribution of the flow magnitude can be dominated by zero values in certain scenes with a still camera.
Such an unbalanced distribution causes the model to prefer generating small motions.
To alleviate this unsatisfactory issue, for each tracking point $p \in \mathcal{P}$, we apply an additional per-pixel weighting factor $w_p$ based on the flow magnitude.
Overall, we train $F$ with the following objective:

\begin{equation}
    \label{eq:flow_loss}    
    \begin{split}
        \mathcal{L}_{F} &= \frac{1}{\|\mathcal{P}\|} \sum_{p \in \mathcal{P}}{ w_p \cdot \| \, d_{a \to b}(p) - \hat d_{a \to b}(p) \, \|_2} \, ,\\
        w_p &= \lambda + \|\,\hat d_{a \to b}(p)\,\|_2 \, / \, \hat{d}_{\text{max}} \, ,
    \end{split}
\end{equation}
where $\hat{d}_{\text{max}}$ is the largest flow magnitude in $\|\,\hat d_{a \to b}(p)\,\|_2$, and $\lambda$ is the minimal loss weight of a pixel.

\subsection{Future-Frame Prediction}
\label{sec:future_frame}
% purpose of future-frame predictor
% given the ... of diffusion model, we formulate the video frame synthesis as a diffusion denoising process
% (unconditional diffusion model formulation)
% motivated by LDM, we apply the conditioning mechanism by concatenating the ...
% to reduce computation, we reduce by ... latent

Next, we generate the future frame $x_{(i+1)}$ based on the current frame $x_i$ and the predicted dense flow map $d_{i \to (i+1)}$.
We formulate the video frame synthesis problem into a sequence of denoising diffusion process~\cite{ddpm} to leverage the expressiveness of diffusion models.
Formally, to synthesize a future frame $x_{(i+1)}$, we first sample a variable $x^{T}_{(i+1)}$ from a Gaussian distribution and then gradually denoise the noisy variable $x^{t}_{(i+1)}$ (where $t = T, ..., 1$) through a UNet with $T$ time steps.
To make the denoising process conditioned on the current frame $x_i$ and the dense flow map $d_{i \to (i+1)}$, we follow LDM~\cite{ldm} and concatenate the noisy variable $x^{t}_{(i+1)}$ with two conditions, $x_i$ and $d_{i \to (i+1)}$, to form the input data of the denoising UNet.
Finally, we take the output $x^{0}_{(i+1)}$ as the predicted future frame $x_{(i+1)}$.

To train the future frame prediction module $G$ with any pair of training video frames $(x_a, x_b)$, we take the current frame $x_a$ and the extracted dense flow map $\hat d_{a \to b}$ as inputs and supervise $G$ by the objective of $\epsilon$ prediction~\cite{ddpm}:

\begin{equation}
    \label{eq:frame_loss}
    % \mathcal{L}_{frame} = \mathbb{E}_{x_b, \epsilon \sim (0,1), t} \left[ \left\| \epsilon - \epsilon_\theta(x_b^t, t, x_a, \hat d_{a \to b}) \right\|^2_2 \right],
    \mathcal{L}_{G} = \mathop{\mathbb{E}}_{x_b, \epsilon \sim (0,1), t} \left\| \, \epsilon - \epsilon_\theta(x_b^t, t, x_a, \hat d_{a \to b}) \, \right\|^2_2 \, ,
\end{equation}
where $\epsilon_\theta$ is the UNet operation and $t$ is uniformly sampled from $\{1, ..., T\}$.

% Hubert: This thing should not be included in the last paragraph of future-frame prediction model, as it involves training flow completion model as well
\subsection{End-to-End Fine-Tuning}
\label{sec:end2end}
%\hubert{
After $F$ and $G$ are separately trained until converging, we end-to-end fine-tune the whole pipeline to reduce the domain gap between the two models and better synergize their interaction.
Specifically, we cascade two models by making $G$ conditioning on the current frame $x_a$ and the predicted dense flow map $d_{a \rightarrow b}$ from $F$, formulating an end-to-end differentiable pipeline $G(x_a, F(x_a, s_{a \to b}))$.
% by taking the predicted dense flow maps $d$ from $F$ as the input conditions of $G$ to synthesize the future frame.
%
% We apply the following objective to fine-tune the whole framework in an end-to-end manner:
We fine-tune the whole pipeline in an end-to-end manner with the following objective:

\begin{equation} 
    \label{eq:all_loss}
    \mathcal{L} = \lambda_{F} \cdot \mathcal{L}_{F} + \lambda_{G} \cdot \mathcal{L}_{G}
\end{equation}
with hyperparameters $\lambda_{F}$ and $\lambda_{G}$.
%}

%% file: 4_experiment.tex
\section{Experiments}
%We validate our framework on three large-scale video datasets, as described in Section~\ref{sec:dataset}, and present the implementation details in Section~\ref{sec:details}.
%
%In Section~\ref{sec:sota}, we quantitatively and qualitatively compare our model with the prior methods on two commonly used benchmarks, while we further examine the limit of our model and conduct an ablation study on MPII Human Pose dataset~\cite{mpii} in Section~\ref{sec:mpii}.

\subsection{Datasets}
\label{sec:dataset}
We validate MCDiff on two general benchmarks and further examine its capability of generating videos with diverse contents and motions on MPII Human Pose~\cite{mpii}.
%test the real-world applicability of our framework on MPII Human Pose~\cite{mpii} dataset, consisting of videos with diverse contexts and motions.

\noindent\textbf{TaiChi-HD}~\cite{taichi} is a collection of 280 in-the-wild TaiChi videos from Youtube.
Each video contains one person and is captured by a fixed or slightly shifting camera viewpoint.
We follow the official training and testing split list to subdivide the dataset into 252 training and 28 testing videos.

\noindent\textbf{Human3.6M}~\cite{human36m} is a human motion dataset with video sequences of 7 actors performing 17 actions (\eg, eating and sitting).
Each video contains only one actor and is captured indoors with a fixed camera viewpoint.
We employ the standard data pre-processing~\cite{SRVP,unsupervised,hierarchical} by center-cropping and downsampling the videos to 6.25 Hz and use actors S1, S5, S6, S7, and S8 (600 videos) for training and actors S9 and S11 (239 videos) for testing.

\noindent\textbf{MPII Human Pose} (MPII)~\cite{mpii} collects 24,987 high-resolution video clips from YouTube, which contain 410 diverse human activities, covering daily activities (\eg, walking) and strenuous exercise (\eg, jumping and skiing).
Each video includes one or multiple humans and is captured under various camera adjustments.
%
%Given the diverse contents and motions, we use this dataset to validate the real-world applicability of our model.
%
%For the dataset pre-processing, 
We remove the videos with scene transition effects or invalid aspect ratios (narrower than $1:1$ or wider than $2:1$) and split the dataset into 19,609 training and 2,000 testing video clips.

\subsection{Implementation Details}
\label{sec:details}
\noindent\textbf{Model Architecture.}
We use LDM-4~\cite{ldm} for both the UNets in the flow completion model $F$ and the future-frame prediction module $G$.
To achieve high-resolution video synthesis on a limited computational resource, we follow LDM by encoding video frames with the spatial size of $256 \times 256$ into latent variables with the size of $64 \times 64$ through a pre-trained VQ-4 autoencoder~\cite{ldm}.
During inference, after the denoising process, we decode latent variables into output videos with the resolution of $256 \times 256$.

\noindent\textbf{Video Dynamics Annotations.}
To track human body keypoints, we first apply HRNet~\cite{hrnet} to frame-wisely estimate the location of 17 keypoints for each person and then connect the person's identity between two adjacent frames by the Hungarian algorithm.
%
%To connect the landmarks among all video frames, we then compute the displacement of each landmark pair between two adjacent frames and perform the Hungarian algorithm to connect the set of landmark pairs with the minimum overall displacement.
%
For the tracking of general points, we perform PIP~\cite{particle} to track a grid of $64 \times 64$ uniformly scattered points for MPII~\cite{mpii} and $16 \times 16$ points for the other two datasets.
%
%Specifically, we sample a grid of points in the middle frame of a video and trace the point forward and reversely for the first and second half of the video to reduce the unsatisfactory tracking results for a long video.
%

\noindent\textbf{Model Training.}
Given a video, we sample a pair of frames $(x_a, x_b)$ with the interval $(b-a) \in [4, 6]$ for training.
As described in Section~\ref{sec:flow_completion}, we downsample the extracted dense flows $\hat d_{a \to b}$ to get the sparse flows $s_{a \to b}$ for training $F$.
Specifically, we randomly sample the flows of two sets of tracking points: (a) 30\% of keypoints for each person and (b) 8 general points for MPII and 4 for the other two.
Note that the sampling probability of a keypoint is correlated to its flow magnitude while the general points are uniformly sampled.
The training of MCDiff includes two steps.
First, we separately train $F$ and $G$ for 400k iterations with a batch size of 40 and a learning rate of 4.5e-6 for $F$ and 7e-5 for $G$ and set $\lambda$ as 0.2.
In the second stage, we fine-tune the whole pipeline with additional 100k iterations with a batch size of 20 and a learning rate of 7e-5.
The weighting factors of $\mathcal{L}$ are $\lambda_F$ = 0.05 and $\lambda_G$ = 1.
We train the model on 8 NVIDIA A100 GPUs.

\begin{table*}[t]
    \centering
    \small
    \mycaption{MCDiff outperforms prior methods on two major benchmarks~\cite{human36m,taichi}}{
    %\hubert{
    We report FVD~\cite{fvd}, LPIPS~\cite{lpips}, SSIM~\cite{ssim}, and PSNR ($\downarrow$ indicates the lower the better, and vice versa).
    Given video synthesis naturally does not have a definitive ground truth for reference-based evaluation, FVD is currently the most reliable metric for assessing visual quality and diversity.
    MCDiff shows substantial performance gain on both datasets under different numbers of input strokes.}
    %}
    \label{table:sota}
    \setlength\tabcolsep{4.5pt} % default value: 6pt
    \begin{tabular}{l||cccccccccccc}
    \hline
    \multirow{3}{*}{Method} & \multicolumn{12}{c}{\textbf{TaiChi-HD}~\cite{taichi}}                                                                           \\ \cline{2-13} 
                            & \multicolumn{4}{c|}{1 strokes}                 & \multicolumn{4}{c|}{5 strokes}                 & \multicolumn{4}{c}{9 strokes} \\
                            & FVD$\downarrow$ & LPIPS$\downarrow$ & SSIM$\uparrow$ & \multicolumn{1}{c|}{PSNR$\uparrow$}
                            & FVD$\downarrow$ & LPIPS$\downarrow$ & SSIM$\uparrow$ & \multicolumn{1}{c|}{PSNR$\uparrow$}
                            & FVD$\downarrow$ & LPIPS$\downarrow$ & SSIM$\uparrow$ & PSNR$\uparrow$  \\ \hline
    II2V~\cite{II2V}        & 194.32          & 0.13          & \textbf{0.75} & \multicolumn{1}{c|}{\textbf{20.53}} & 217.41          & 0.16          & 0.72          & \multicolumn{1}{c|}{20.01}          & 258.60          & 0.14          & 0.66          & 18.97          \\
    iPOKE~\cite{ipoke}      & 156.97          & \textbf{0.11} & 0.74          & \multicolumn{1}{c|}{20.42}          & 149.82          & \textbf{0.11} & \textbf{0.76} & \multicolumn{1}{c|}{20.31}          & 175.48          & 0.14          & 0.71          & 20.10          \\ \hline
    MCDiff (Ours)           & \textbf{142.57} & \textbf{0.11} & 0.72          & \multicolumn{1}{c|}{20.51}          & \textbf{126.69} & \textbf{0.11} & 0.75          & \multicolumn{1}{c|}{\textbf{20.80}} & \textbf{113.12} & \textbf{0.10} & \textbf{0.77} & \textbf{21.45} \\ \hline
    \end{tabular}
    
    \vspace{+2mm}
    
    \begin{tabular}{l||cccccccccccc}
    \hline
    \multirow{3}{*}{Method} & \multicolumn{12}{c}{\textbf{Human3.6M}~\cite{human36m}}                                                                         \\ \cline{2-13} 
                            & \multicolumn{4}{c|}{1 strokes}                 & \multicolumn{4}{c|}{5 strokes}                 & \multicolumn{4}{c}{9 strokes} \\
                            & FVD$\downarrow$ & LPIPS$\downarrow$ & SSIM$\uparrow$ & \multicolumn{1}{c|}{PSNR$\uparrow$}
                            & FVD$\downarrow$ & LPIPS$\downarrow$ & SSIM$\uparrow$ & \multicolumn{1}{c|}{PSNR$\uparrow$}
                            & FVD$\downarrow$ & LPIPS$\downarrow$ & SSIM$\uparrow$ & PSNR$\uparrow$  \\ \hline
    II2V~\cite{II2V}        & 130.53          & 0.08          & 0.90          & \multicolumn{1}{c|}{\textbf{22.73}} & 143.27          & 0.11          & 0.88          & \multicolumn{1}{c|}{21.04}          & 173.60          & 0.15          & 0.81          & 20.23          \\
    iPOKE~\cite{ipoke}      & 118.02          & \textbf{0.06} & \textbf{0.91} & \multicolumn{1}{c|}{21.88}          & 117.57          & \textbf{0.05} & 0.91          & \multicolumn{1}{c|}{22.21}          & 137.88          & 0.09          & 0.84          & 21.57          \\ \hline
    MCDiff (Ours)           & \textbf{117.60} & 0.09          & \textbf{0.91} & \multicolumn{1}{c|}{22.07}          & \textbf{114.82} & 0.06          & \textbf{0.93} & \multicolumn{1}{c|}{\textbf{23.51}} & \textbf{111.38} & \textbf{0.05} & \textbf{0.94} & \textbf{23.73} \\ \hline
    \end{tabular}
\end{table*}

\begin{figure*}[t]
    \centering
    \includegraphics[width=\linewidth]{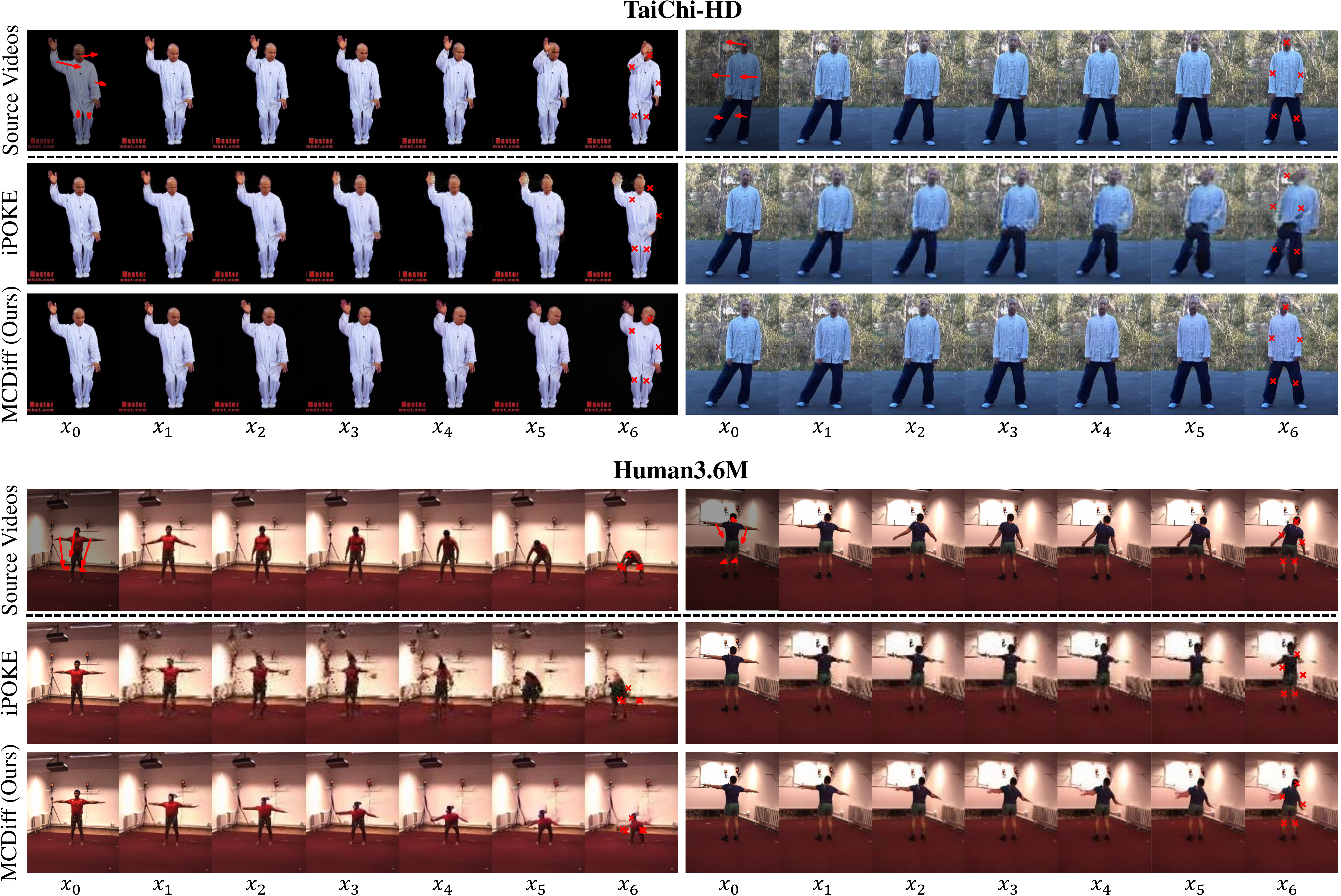}
    \mycaption{MCDiff achieves better visual quality and controllability compared to iPOKE~\cite{ipoke} on two benchmarks~\cite{human36m,taichi}}{
        We input iPOKE~\cite{ipoke} and MCDiff with the same start frames (the leftmost column of each sequence) and motions of 5 human keypoints ({\color{red} red} arrows on the nose, elbows, and knees), which are sampled from the testing videos (the top row of each sequence).
        The target locations of the keypoints are marked at the end frames ({\color{red} red} crosses at the rightmost column of each sequence).
        MCDiff is able to synthesize the videos with better quality while more faithfully following the motions specified by the strokes.
    }
    \label{fig:ipoke}
\end{figure*}

\subsection{Comparisons with Prior Methods}
\label{sec:sota}
In this section, we compare MCDiff against the state-of-the-art stroke-guided controllable video synthesis methods, II2V~\cite{II2V} and iPOKE~\cite{ipoke}.
To achieve a fair comparison, we input the same start frames and strokes extracted from the testing videos of TaiChi-HD~\cite{taichi} and Human3.6M~\cite{human36m}.
Specifically, we take the displacements of the human body keypoints (\ie, the shortest path from the start to the end location) as the input strokes.
As MCDiff takes momentary motion at each time step as input, to avoid the leakage of testing data information, we linearly and uniformly split the displacements of keypoints into multiple segments as the inputs.
For all competitors, we use the official codes and the provided pre-trained models for inference.

\noindent\textbf{Quantitative Evaluation.}
% What is the experiment setting
% Where and what are the results
We first quantitatively evaluate the synthesis quality by reporting the scores of Fr\'echet Video Distance (FVD)~\cite{fvd}, which is responsive to visual quality, content diversity, and temporal coherence.
We compute FVD based on the video length of 10.
Following the standard practice, we also apply frame-wise evaluation metrics, including LPIPS~\cite{lpips}, SSIM~\cite{ssim}, and PSNR, while pointing out that video synthesis naturally does not have a definitive ground truth for reference-based evaluations.
%
%To compare the for finer-grained control when users want to assign an accurate motion with more strokes.
%
The results are reported in Table~\ref{table:sota}.

Leveraging the expressiveness of the diffusion models, our model sets a new state-of-the-art synthesis quality in stroke-guided controllable video synthesis.
Moreover, the prior approaches usually fall short of maintaining the synthesis quality with increasing numbers of input strokes.
By comparison, MCDiff performs consistently with different numbers of strokes, inferring the flexibility of MCDiff to handle the case when a user is eager to achieve finer motion control by specifying more input strokes.

\noindent\textbf{Qualitative Comparisons.}
% What is the experiment setting
% Where and what are the results
We then visualize the synthetic videos from MCDiff and iPOKE in Figure~\ref{fig:ipoke}.
For both models, we input the same start frames and strokes, including the displacements of the nose, elbows, and knees, sampled from the testing set of TaiChi-HD and Human3.6M.

MCDiff outperforms iPOKE in terms of visual quality and controllability.
Specifically, iPOKE usually generates videos with unnatural distortion on the human body and fails to follow the motion instructed by the input strokes.
For example, in the bottom-left sample of Figure~\ref{fig:ipoke}, the end frame generated by iPOKE does not preserve a reasonable human body shape, and the human keypoints with the motion control do not faithfully move to the target destinations.
In contrast, MCDiff is able to achieve a better visual quality while more precisely matching the motion conditions.
In Appendix~\ref{sec:control}, we further provide the quantitative experiment on motion controllability and demonstrate the generality of this observation.

\begin{table}
    \centering
    %\small
    %\vspace{-1.5mm}
    \mycaption{Ablation analysis of Flow Completion Model on MPII Human Pose~\cite{mpii}}{
        We report FVD~\cite{fvd}, LPIPS~\cite{lpips}, SSIM~\cite{ssim}, and PSNR.
        The numbers show the effectiveness of the two-stage model design.
    }
    \label{table:ablation}
    %\vspace{-1.5mm}
    \begin{tabular}{l|cccc}
    \hline
    \multirow{2}{*}{Method} & \multicolumn{4}{c}{\textbf{MPII Human Pose}~\cite{mpii}} \\ \cline{2-5} 
                            & FVD$\downarrow$ & LPIPS$\downarrow$ & SSIM$\uparrow$ & PSNR$\uparrow$ \\ \hline
    Ours (w/o $F$)          & 273.86          & 0.18              & 0.63           & 18.34          \\ \hline
    Ours (Full)             & \textbf{194.30} & \textbf{0.14}     & \textbf{0.69}  & \textbf{19.52} \\ \hline
    \end{tabular}
\end{table}

\begin{figure}[t]
    \centering
    \includegraphics[width=\linewidth]{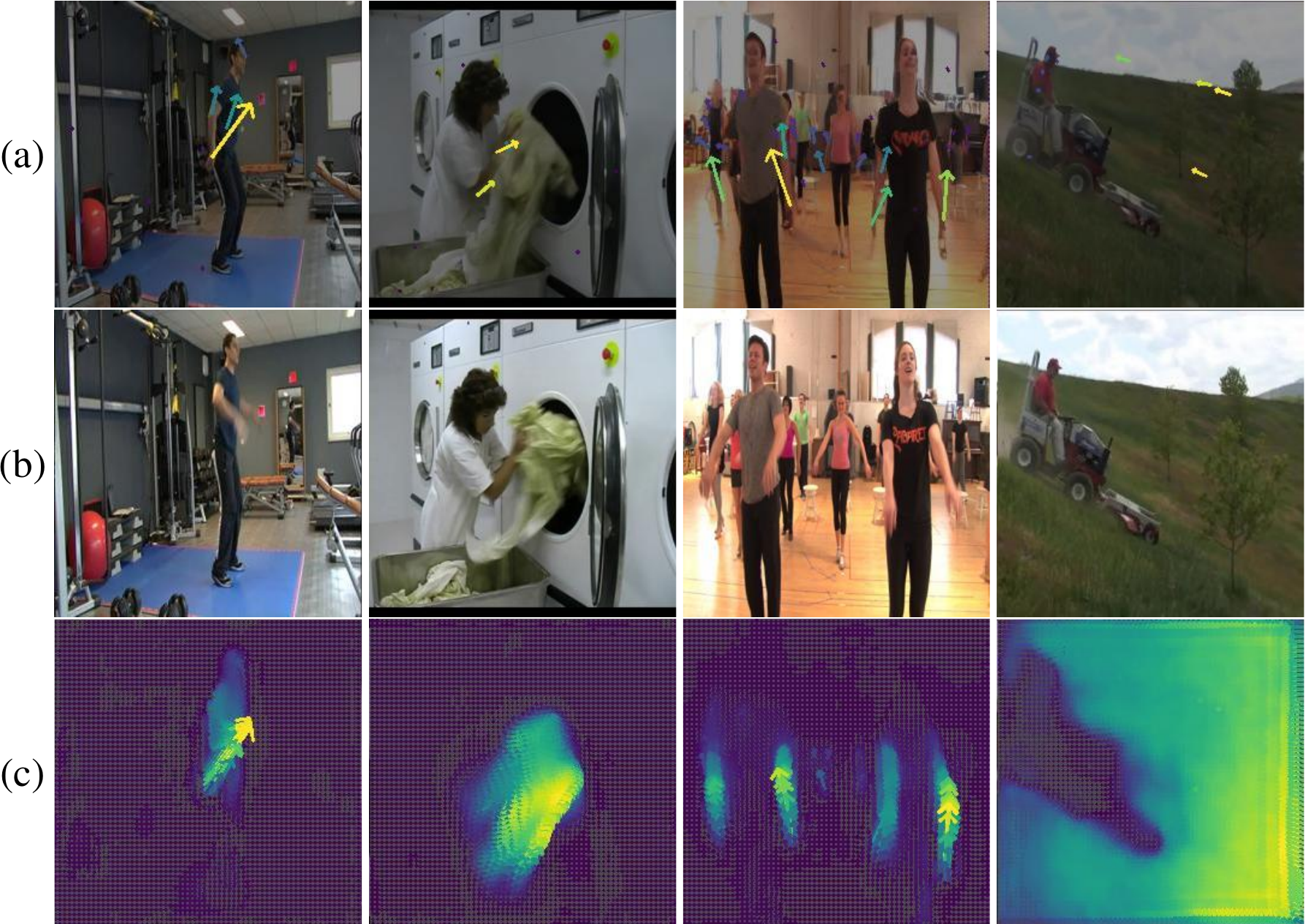}
    \mycaption{Qualitative Results of Flow Completion Model on MPII Human Pose~\cite{mpii}}{
        %\hubert{
        From top to bottom, we show (a) the input frame with the strokes, (b) the next frame in the real video, and (c) our predicted flows.
        Our flow completion model can predict high-quality flows based on the semantic understanding of the video frame and the sparse motion control.
        %We showcase four types of difficult motion prediction and demonstrate that our model can handle these diverse situations and produce high-quality flows.
        %
        %From the left to right columns are activities with large motion, the interaction between objects (\eg, hands and clothes,) multiple objects movements, and camera viewpoint shifting. 
        %}
    }
    \label{fig:flow_completion}
\end{figure}

\begin{figure*}[t]
    \centering
    \includegraphics[width=\linewidth]{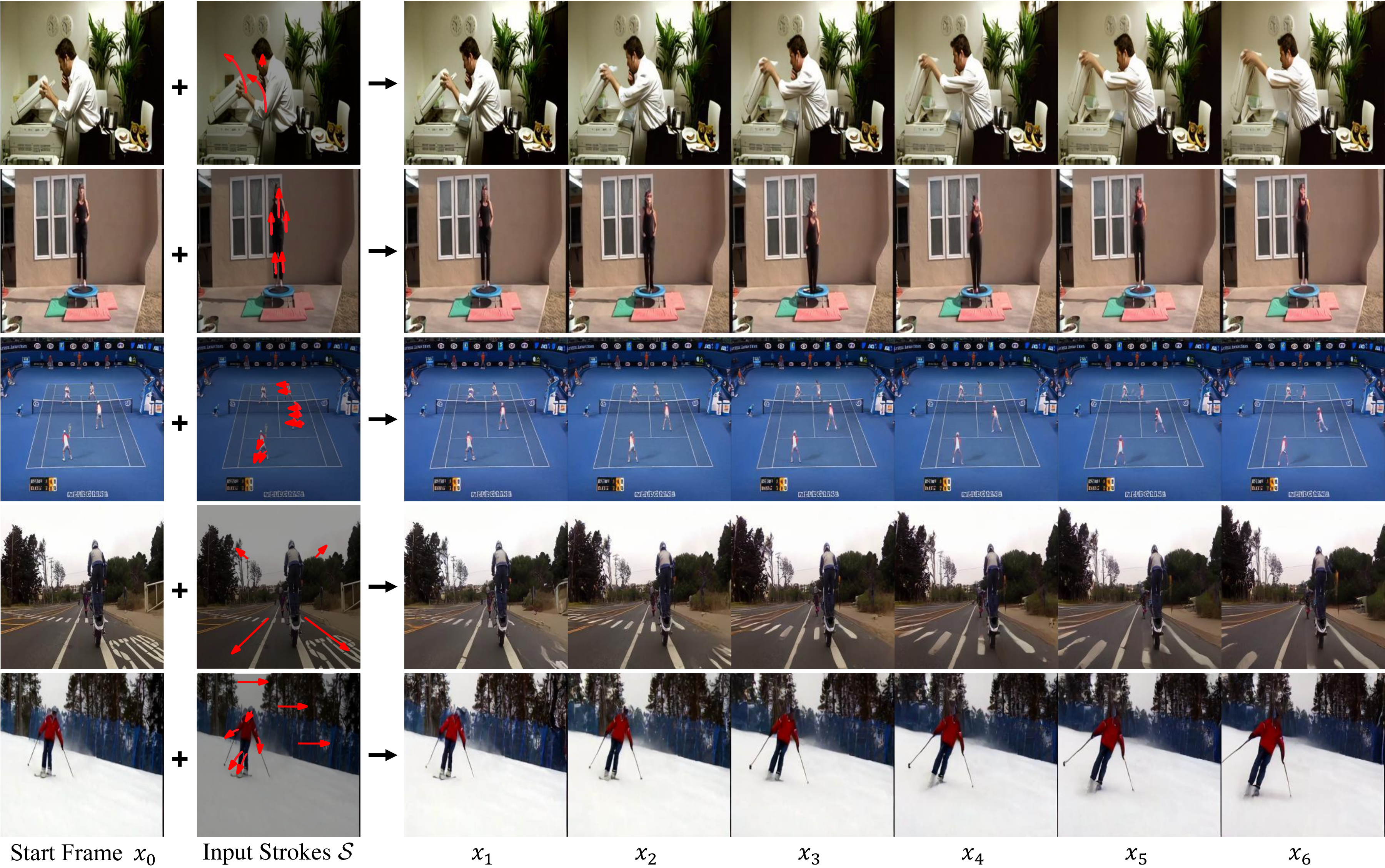}
    %\vspace{-3.8mm}
    \mycaption{MCDiff synthesizes high-quality videos with diverse contents and motions}{
        We input MCDiff with the start frames sampled from the testing videos of MPII Human Pose~\cite{mpii} and the strokes manually specified by humans. 
        Under diverse conditions and activities, MCDiff can synthesize high-quality and temporally consistent videos while attending to the input conditions.  
        Noticeably, in the bottom two rows, the model is aware of the strokes assigned on the background scene indicating the camera adjustment.
        %Notice the model is aware of certain strokes indicating the camera adjustment in the last two rows.
        % 
        We show more visual results in Appendix~\ref{sec:visualization}.
    }
    \label{fig:mpii}
    %\vspace{-1.5mm}
\end{figure*} 

\subsection{Failure of Single-Stage Framework}
\label{sec:single-stage}
% What is the experiment setting
%\ts{I'll revise this paragraph later.}
Next, we conduct the ablation study to justify the necessity of the flow completion model $F$ on MPII~\cite{mpii}.
%compare the models with and without the flow completion model $F$
%
For the model without $F$, we directly use the sparse flow maps as inputs of $G$.
%
%We train and test both models on the training and testing videos of MPII, while the rest of the experiment setting is the same as Section~\ref{sec:sota},
We separately train two models on MPII and remain other experiment settings the same as in Section~\ref{sec:sota}.
The results are reported in Table~\ref{table:ablation}.
The full model with the two-stage design outperforms the single-stage counterpart with a clear performance gap.
Such results infer that synthesizing a future frame from a dense flow map can better tackle the ambiguity and difficulty of sparse motion inputs and benefit the procedure of video frame synthesis.

\subsection{Synthesis with Diverse Contents and Motions}
\label{sec:mpii}
In this section, we further evaluate the capability of the flow completion model $F$ and MCDiff to synthesize videos with diverse contents and motions.

\noindent\textbf{Flow Completion Model.}
We first input $F$ with the video frames and sparse strokes sampled from the testing videos of MPII~\cite{mpii} and visualize the results in Figure~\ref{fig:flow_completion}.
From the left to the right columns, we showcase four types of difficult motion predictions, including the activity with large motion, the interaction between objects (\eg, hands and clothes), multiple objects movements, and camera viewpoint shifting.
We demonstrate that our model can handle these diverse situations and predict high-quality dense flow with regard to the semantic understanding of the input image and sparse strokes.
%
%For example, the leftmost sample of Figure~\ref{fig:flow_completion} shows that $F$ can learn to propagate the sparse motions on the elbow and wrist to the entire arm with reasonable magnitudes and directions, representing a human raising hand.
%
For example, the second left sample of Figure~\ref{fig:flow_completion} shows that $F$ can recognize the action of a human grabbing clothes and predict the coherent motion between both the hands and clothes.
Moreover, in the rightmost sample of Figure~\ref{fig:flow_completion}, we observe that $F$ discovers the strokes in the background scene as the control of the camera viewpoint, and hence, generates the consistent motions on the backgrounds.
Such results show the capability of $F$ to well tackle the difficulty of sparse motion inputs and achieve more flexible motion control with camera adjustments.

\noindent\textbf{Video Synthesis.}
Next, we test the video synthesis capability of MCDiff pre-trained on MPII.
The start frames are sampled from the testing videos of MPII, and the strokes are manually specified by the users.
In Figure~\ref{fig:mpii}, we showcase five samples representing different types of content and motion conditions.
From the top to the bottom row, we respectively show the samples with the coherent movements of both foreground and background objects, huge foreground object movement, multiple foreground objects movements, significant background motion, and both foreground motion and camera viewpoint shifting. 
These results exhibit that MCDiff can leverage the expressiveness of the diffusion model and synthesize high-quality videos under diverse content and motion conditions.
Moreover, in the bottom row, we find that MCDiff learns to identify the strokes on the foreground subject as the object movements, while strokes on the background scene represent camera viewpoint changes.
Such behavior enables the users to achieve more flexible control and synthesize more realistic camera trajectories for video synthesis and editing.

%% file: 5_conclusion.tex
\section{Conclusion and Limitations}

%\hubert{
We present MCDiff, a powerful diffusion model framework for controllable video synthesis, achieving high-quality and high-fidelity video synthesis with an intuitive and fine-grained user control interface.
The method shows promising results on three large-scale video synthesis datasets with a wide range of human activities.
%}

%\hubert{
Despite showing impressive results, our framework is still bounded by a few limitations.
First, since the flow completion model is learned in a data-driven manner, it is difficult to fulfill edits that are well beyond the training distributions.
For instance, in Figure~\ref{fig:teaser}, moving the score bug from the top-left to the middle of the screen, or interpreting the flows in the background as all the storage collapses down from the shelves.
Learning a generative model that can handle novel compositions of objects~\cite{azadi2020compositional,dhamo2020semantic} remains challenging, and we put learning such a generalist model as an important future direction.
%}

%\hubert{
Second, the training flows are extracted with the methods based on pattern recognition, as common internet videos do not contain special sensory information about the actual motion field. 
Since these methods are designed to solve the relationship of image patterns across video frames, they typically fail on videos with texture-less surfaces or visual illusion (\eg, the barber-pole illusion) cases.
Therefore, our model inherits such limitations from the training data and is unlikely to operate in these special cases.
Future efforts in building video datasets with physically grounded motion fields using special sensors or simulations can help address these difficult and rare cases.
%}

%% file: 6_appendix.tex
%\section*{Supplementary Material}
In the appendix, we first quantitatively compare the motion controllability of our model with prior methods in Section~\ref{sec:control} and provide more visualization results on MPII Human Pose~\cite{mpii} dataset in Section~\ref{sec:visualization}.
%
%In the supplementary martial, we also provide the videos (.mp4 files) for each visualized video result in the main paper and the appendix.

\vspace{+2mm}
\section{Comparison of Motion Controllability}
\label{sec:control}
In this section, we compare the motion controllability of MCDiff against II2V~\cite{II2V} and iPOKE~\cite{ipoke}, the state-of-the-art stroke-guided controllable video synthesis methods.
Specifically, in Section~\ref{sec:sota}, we quantitatively show that MCDiff is able to generate video sequences with the motions more faithfully adhering to the input strokes.
The observation infers that MCDiff offers a user better controllability to achieve the anticipated motion within the synthetic video.
Here, we further provide the quantitative experiments for this finding.

Following the experiment setting in Section~\ref{sec:sota}, we use the same input start frames and strokes for all methods and test the models with different numbers of input strokes.
The start frames and strokes are sampled from the testing videos of TaiChi-HD~\cite{taichi} and Human3.6M~\cite{human36m}.
To be more detailed, we test the models with 1, 5, or 9 input strokes.
For the test cases of 5 strokes, we sample the displacements of the nose, elbows, and knees as the input strokes, while additionally sampling the displacements of the wrists and ankles for the test cases of 9 strokes.
For the cases of 1 stroke, we randomly sample the displacement of the nose, elbow, or knee.

To quantitatively measure the motion controllability of a video synthesis framework, or how faithfully a framework follows the instruction of the input strokes to synthesize a video, we perform human pose estimation, HRNet~\cite{hrnet}, on the end video frame and compute the Displacement Error (DE) for each human keypoint controlled by an input stroke.
DE is the Euclidean distance between the target location and the actual ending-up location of a human keypoint.
We compute DE based on the videos with the resolution of $128 \times 128$ and average the DE scores among all controlled keypoints, as Average Displacement Error (ADE).
Considering that the human pose estimation algorithm might be hard to detect the human keypoints on a low-quality video frame, we additionally report the recall rate to indicate the percentage of the retrieved keypoints.
The recall rate is computed by using the threshold of $0.2$ and viewing the keypoints with confidence scores less than $0.2$ as the miss detections.
We report the results in Table~\ref{table:control}.

Generally, MCDiff achieves lower ADE and a higher recall rate compared to the previous methods.
The performance gap is more significant when the number of input strokes increases.
The results demonstrate that MCDiff is able to synthesize a video more precisely matching the input strokes and can enable more flexible and accurate motion control when a user is eager to specify more input strokes to achieve finer control.
Additionally, a higher recall rate indicates our method synthesizes video frames with more reasonable human body appearances and structures, resulting in less miss detection rate from the human pose estimation algorithm.

\begin{table}[h]
    \centering
    \small
    \mycaption{MCDiff achieves superior motion controllability on two major benchmarks~\cite{human36m,taichi}}{
    %\hubert{
    We report the Average Displacement Error (AVD) and the recall rate (Recall).
    MCDiff outperforms prior methods with a substantial performance gap on both datasets under different numbers of input strokes.}
    %}
    \label{table:control}
    \setlength\tabcolsep{5.0pt}
    \begin{tabular}{l||cccccc||cccccc}
    \hline
    \multirow{3}{*}{Method} & \multicolumn{6}{c||}{\textbf{TaiChi-HD}~\cite{taichi}} & \multicolumn{6}{c}{\textbf{Human3.6M}~\cite{human36m}}                  \\ \cline{2-13} 
                            & \multicolumn{2}{c|}{1 strokes} & \multicolumn{2}{c|}{5 strokes} & \multicolumn{2}{c||}{9 strokes} & \multicolumn{2}{c|}{1 strokes} & \multicolumn{2}{c|}{5 strokes} & \multicolumn{2}{c}{9 strokes} \\
                            & ADE$\downarrow$ & \multicolumn{1}{c|}{Recall$\uparrow$} & ADE$\downarrow$ & \multicolumn{1}{c|}{Recall$\uparrow$} & ADE$\downarrow$ & \multicolumn{1}{c||}{Recall$\uparrow$}  
                            & ADE$\downarrow$ & \multicolumn{1}{c|}{Recall$\uparrow$} & ADE$\downarrow$ & \multicolumn{1}{c|}{Recall$\uparrow$} & ADE$\downarrow$ & \multicolumn{1}{c}{Recall$\uparrow$} \\ \hline
    II2V~\cite{II2V}        & 4.17          & \multicolumn{1}{c|}{0.99}          & 7.51          & \multicolumn{1}{c|}{0.97}          & 11.36         & \multicolumn{1}{c||}{0.94}          
                            & 2.54          & \multicolumn{1}{c|}{0.94}          & 3.18          & \multicolumn{1}{c|}{0.88}          & 4.92          & \multicolumn{1}{c}{0.81} \\ \hline
    iPOKE~\cite{ipoke}      & \textbf{2.63} & \multicolumn{1}{c|}{\textbf{1.00}} & 5.09          & \multicolumn{1}{c|}{0.99}          & 8.94          & \multicolumn{1}{c||}{0.96}          
                            & 1.80          & \multicolumn{1}{c|}{0.97}          & 2.53          & \multicolumn{1}{c|}{0.93}          & 3.74          & \multicolumn{1}{c}{0.89} \\ \hline
    MCDiff (Ours)           & 2.77          & \multicolumn{1}{c|}{0.99}          & \textbf{2.72} & \multicolumn{1}{c|}{\textbf{1.00}} & \textbf{2.90} & \multicolumn{1}{c||}{\textbf{1.00}} 
                            & \textbf{1.64} & \multicolumn{1}{c|}{\textbf{0.98}} & \textbf{1.73} & \multicolumn{1}{c|}{\textbf{0.98}} & \textbf{1.86} & \multicolumn{1}{c}{\textbf{0.96}} \\ \hline
    \end{tabular}
\end{table}

\section{Additional Visualizations on MPII Human Pose }
\label{sec:visualization}
In this section, we visualize both the dense flow maps and video frames predicted from the flow completion model $F$ and the future-frame prediction module $G$.
We show the input start frame $x_0$ and strokes $\mathcal{S}$ on the left.
And on the right, we plot the momentary strokes $s$, dense flow map $d$, and the synthesized video frames, arranged from top to bottom.
The start frames are sampled from the testing videos of MPII Human Pose~\cite{mpii}, while the strokes are manually specified by humans.
\clearpage

\begin{figure}[ht]
\centering
    \includegraphics[width=\linewidth]{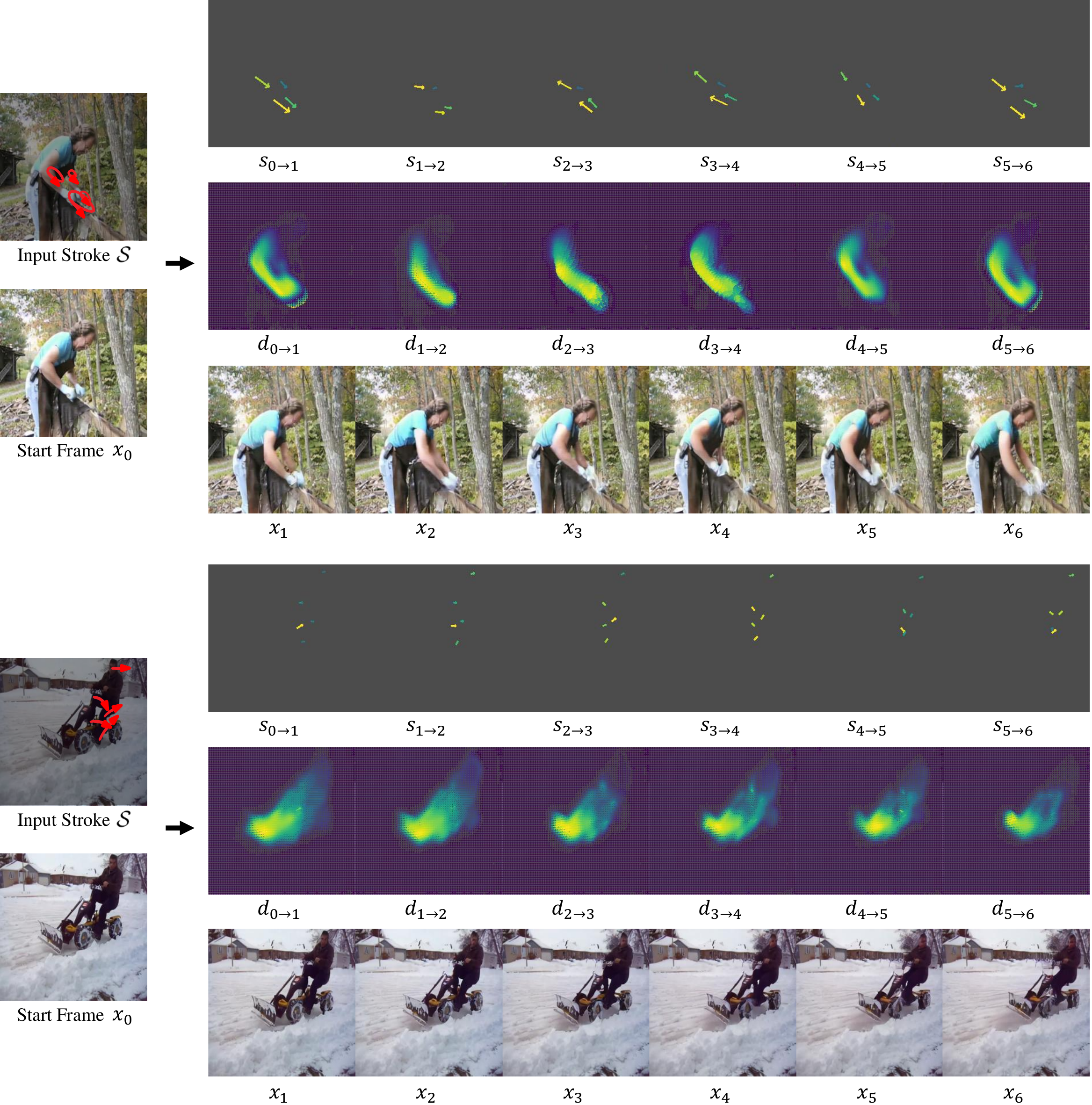}
    \captionsetup{labelformat=empty}
\end{figure}
\clearpage

\begin{figure}[ht]
\centering
    \includegraphics[width=\linewidth]{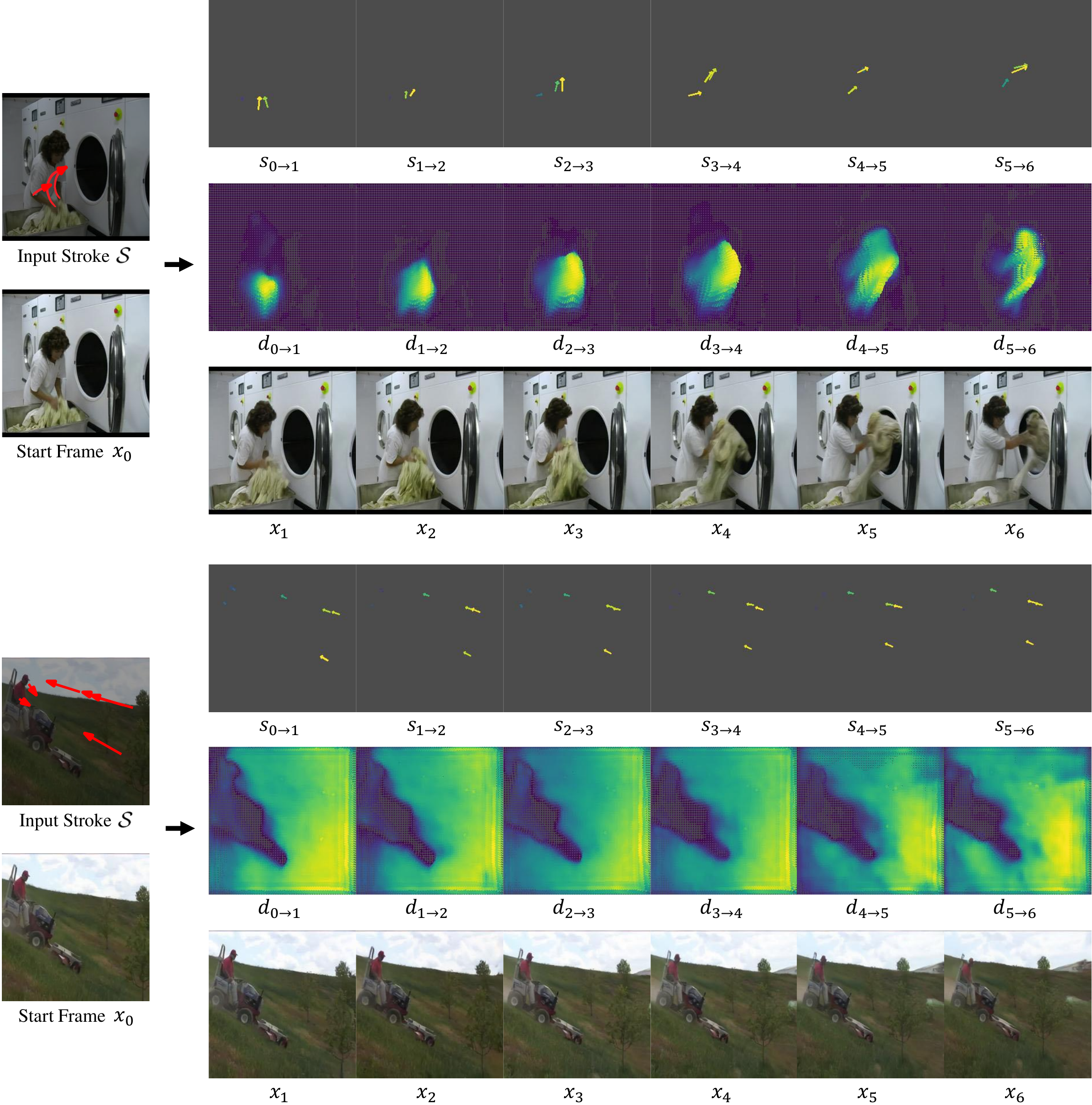}
    \captionsetup{labelformat=empty}
\end{figure}

%% file: main.bbl
\begin{thebibliography}{10}\itemsep=-1pt

\bibitem{mpii}
Mykhaylo Andriluka, Leonid Pishchulin, Peter Gehler, and Bernt Schiele.
\newblock 2d human pose estimation: New benchmark and state of the art
  analysis.
\newblock In {\em IEEE Conference on Computer Vision and Pattern Recognition},
  2014.

\bibitem{clicktomove}
Pierfrancesco Ardino, Marco De~Nadai, Bruno Lepri, Elisa Ricci, and
  St{\'e}phane Lathuili{\`e}re.
\newblock Click to move: Controlling video generation with sparse motion.
\newblock In {\em IEEE International Conference on Computer Vision}, 2021.

\bibitem{azadi2020compositional}
Samaneh Azadi, Deepak Pathak, Sayna Ebrahimi, and Trevor Darrell.
\newblock Compositional gan: Learning image-conditional binary composition.
\newblock {\em International Journal of Computer Vision}, 2020.

\bibitem{ediffi}
Yogesh Balaji, Seungjun Nah, Xun Huang, Arash Vahdat, Jiaming Song, Karsten
  Kreis, Miika Aittala, Timo Aila, Samuli Laine, Bryan Catanzaro, et~al.
\newblock ediffi: Text-to-image diffusion models with an ensemble of expert
  denoisers.
\newblock {\em arXiv preprint arXiv:2211.01324}, 2022.

\bibitem{text2live}
Omer Bar-Tal, Dolev Ofri-Amar, Rafail Fridman, Yoni Kasten, and Tali Dekel.
\newblock Text2live: Text-driven layered image and video editing.
\newblock In {\em European Conference on Computer Vision}, 2022.

\bibitem{ipoke}
Andreas Blattmann, Timo Milbich, Michael Dorkenwald, and Bj{\"o}rn Ommer.
\newblock ipoke: Poking a still image for controlled stochastic video
  synthesis.
\newblock In {\em IEEE International Conference on Computer Vision}, 2021.

\bibitem{II2V}
Andreas Blattmann, Timo Milbich, Michael Dorkenwald, and Bjorn Ommer.
\newblock Understanding object dynamics for interactive image-to-video
  synthesis.
\newblock In {\em IEEE Conference on Computer Vision and Pattern Recognition},
  2021.

\bibitem{GRU}
Junyoung Chung, Caglar Gulcehre, KyungHyun Cho, and Yoshua Bengio.
\newblock Empirical evaluation of gated recurrent neural networks on sequence
  modeling.
\newblock {\em arXiv preprint arXiv:1412.3555}, 2014.

\bibitem{pix2pix3D}
Kangle Deng, Gengshan Yang, Deva Ramanan, and Jun-Yan Zhu.
\newblock 3d-aware conditional image synthesis.
\newblock {\em arXiv preprint arXiv:2302.08509}, 2023.

\bibitem{bert}
Jacob Devlin, Ming-Wei Chang, Kenton Lee, and Kristina Toutanova.
\newblock Bert: Pre-training of deep bidirectional transformers for language
  understanding.
\newblock {\em arXiv preprint arXiv:1810.04805}, 2018.

\bibitem{dhamo2020semantic}
Helisa Dhamo, Azade Farshad, Iro Laina, Nassir Navab, Gregory~D Hager, Federico
  Tombari, and Christian Rupprecht.
\newblock Semantic image manipulation using scene graphs.
\newblock In {\em IEEE Conference on Computer Vision and Pattern Recognition},
  2020.

\bibitem{esser2023structure}
Patrick Esser, Johnathan Chiu, Parmida Atighehchian, Jonathan Granskog, and
  Anastasis Germanidis.
\newblock Structure and content-guided video synthesis with diffusion models.
\newblock {\em arXiv preprint arXiv:2302.03011}, 2023.

\bibitem{SRVP}
Jean-Yves Franceschi, Edouard Delasalles, Micka{\"e}l Chen, Sylvain Lamprier,
  and Patrick Gallinari.
\newblock Stochastic latent residual video prediction.
\newblock In {\em International Conference on Machine Learning}, 2020.

\bibitem{scenescape}
Rafail Fridman, Amit Abecasis, Yoni Kasten, and Tali Dekel.
\newblock Scenescape: Text-driven consistent scene generation.
\newblock {\em arXiv preprint arXiv:2302.01133}, 2023.

\bibitem{controllable}
Zekun Hao, Xun Huang, and Serge Belongie.
\newblock Controllable video generation with sparse trajectories.
\newblock In {\em IEEE Conference on Computer Vision and Pattern Recognition},
  2018.

\bibitem{particle}
Adam~W Harley, Zhaoyuan Fang, and Katerina Fragkiadaki.
\newblock Particle video revisited: Tracking through occlusions using point
  trajectories.
\newblock In {\em European Conference on Computer Vision}, 2022.

\bibitem{harvey2022flexible}
William Harvey, Saeid Naderiparizi, Vaden Masrani, Christian Weilbach, and
  Frank Wood.
\newblock Flexible diffusion modeling of long videos.
\newblock {\em arXiv preprint arXiv:2205.11495}, 2022.

\bibitem{MAE}
Kaiming He, Xinlei Chen, Saining Xie, Yanghao Li, Piotr Doll{\'a}r, and Ross
  Girshick.
\newblock Masked autoencoders are scalable vision learners.
\newblock In {\em IEEE Conference on Computer Vision and Pattern Recognition},
  2022.

\bibitem{imagenvideo}
Jonathan Ho, William Chan, Chitwan Saharia, Jay Whang, Ruiqi Gao, Alexey
  Gritsenko, Diederik~P Kingma, Ben Poole, Mohammad Norouzi, David~J Fleet,
  et~al.
\newblock Imagen video: High definition video generation with diffusion models.
\newblock {\em arXiv preprint arXiv:2210.02303}, 2022.

\bibitem{ddpm}
Jonathan Ho, Ajay Jain, and Pieter Abbeel.
\newblock Denoising diffusion probabilistic models.
\newblock {\em Advances in Neural Information Processing Systems}, 2020.

\bibitem{VDM}
Jonathan Ho, Tim Salimans, Alexey~A. Gritsenko, William Chan, Mohammad Norouzi,
  and David~J. Fleet.
\newblock Video diffusion models.
\newblock In {\em Advances in Neural Information Processing Systems}, 2022.

\bibitem{hoppe2022diffusion}
Tobias H{\"o}ppe, Arash Mehrjou, Stefan Bauer, Didrik Nielsen, and Andrea
  Dittadi.
\newblock Diffusion models for video prediction and infilling.
\newblock {\em Transactions on Machine Learning Research}, 2022.

\bibitem{human36m}
Catalin Ionescu, Dragos Papava, Vlad Olaru, and Cristian Sminchisescu.
\newblock Human3.6m: Large scale datasets and predictive methods for 3d human
  sensing in natural environments.
\newblock {\em IEEE Transactions on Pattern Analysis and Machine Intelligence},
  2013.

\bibitem{kim2022diffusion}
Gyeongman Kim, Hajin Shim, Hyunsu Kim, Yunjey Choi, Junho Kim, and Eunho Yang.
\newblock Diffusion video autoencoders: Toward temporally consistent face video
  editing via disentangled video encoding.
\newblock {\em arXiv preprint arXiv:2212.02802}, 2022.

\bibitem{customdiffusion}
Nupur Kumari, Bingliang Zhang, Richard Zhang, Eli Shechtman, and Jun-Yan Zhu.
\newblock Multi-concept customization of text-to-image diffusion.
\newblock {\em arXiv preprint arXiv:2212.04488}, 2022.

\bibitem{magic3D}
Chen-Hsuan Lin, Jun Gao, Luming Tang, Towaki Takikawa, Xiaohui Zeng, Xun Huang,
  Karsten Kreis, Sanja Fidler, Ming-Yu Liu, and Tsung-Yi Lin.
\newblock Magic3d: High-resolution text-to-3d content creation.
\newblock {\em arXiv preprint arXiv:2211.10440}, 2022.

\bibitem{realfusion}
Luke Melas-Kyriazi, Christian Rupprecht, Iro Laina, and Andrea Vedaldi.
\newblock Realfusion: 360° reconstruction of any object from a single image.
\newblock {\em arXiv preprint arXiv:2302.10663}, 2023.

\bibitem{playableenv}
Willi Menapace, St{\'e}phane Lathuiliere, Aliaksandr Siarohin, Christian
  Theobalt, Sergey Tulyakov, Vladislav Golyanik, and Elisa Ricci.
\newblock Playable environments: Video manipulation in space and time.
\newblock In {\em Proceedings of the IEEE/CVF Conference on Computer Vision and
  Pattern Recognition}, pages 3584--3593, 2022.

\bibitem{playablevideo}
Willi Menapace, Stephane Lathuiliere, Sergey Tulyakov, Aliaksandr Siarohin, and
  Elisa Ricci.
\newblock Playable video generation.
\newblock In {\em IEEE Conference on Computer Vision and Pattern Recognition},
  2021.

\bibitem{sdedit}
Chenlin Meng, Yang Song, Jiaming Song, Jiajun Wu, Jun-Yan Zhu, and Stefano
  Ermon.
\newblock Sdedit: Image synthesis and editing with stochastic differential
  equations.
\newblock In {\em International Conference on Learning Representations}, 2022.

\bibitem{unsupervised}
Matthias Minderer, Chen Sun, Ruben Villegas, Forrester Cole, Kevin~P Murphy,
  and Honglak Lee.
\newblock Unsupervised learning of object structure and dynamics from videos.
\newblock {\em Advances in Neural Information Processing Systems}, 2019.

\bibitem{dreamix}
Eyal Molad, Eliahu Horwitz, Dani Valevski, Alex~Rav Acha, Yossi Matias, Yael
  Pritch, Yaniv Leviathan, and Yedid Hoshen.
\newblock Dreamix: Video diffusion models are general video editors.
\newblock {\em arXiv preprint arXiv:2302.01329}, 2023.

\bibitem{sinfusion}
Yaniv Nikankin, Niv Haim, and Michal Irani.
\newblock Sinfusion: Training diffusion models on a single image or video.
\newblock {\em arXiv preprint arXiv:2211.11743}, 2022.

\bibitem{dreamfusion}
Ben Poole, Ajay Jain, Jonathan~T. Barron, and Ben Mildenhall.
\newblock Dreamfusion: Text-to-3d using 2d diffusion.
\newblock In {\em International Conference on Learning Representations}, 2022.

\bibitem{dalle2}
Aditya Ramesh, Prafulla Dhariwal, Alex Nichol, Casey Chu, and Mark Chen.
\newblock Hierarchical text-conditional image generation with clip latents.
\newblock {\em arXiv preprint arXiv:2204.06125}, 2022.

\bibitem{ldm}
Robin Rombach, Andreas Blattmann, Dominik Lorenz, Patrick Esser, and Bj{\"o}rn
  Ommer.
\newblock High-resolution image synthesis with latent diffusion models.
\newblock In {\em IEEE Conference on Computer Vision and Pattern Recognition},
  2022.

\bibitem{dreambooth}
Nataniel Ruiz, Yuanzhen Li, Varun Jampani, Yael Pritch, Michael Rubinstein, and
  Kfir Aberman.
\newblock Dreambooth: Fine tuning text-to-image diffusion models for
  subject-driven generation.
\newblock {\em arXiv preprint arxiv:2208.12242}, 2022.

\bibitem{imagen}
Chitwan Saharia, William Chan, Saurabh Saxena, Lala Li, Jay Whang, Emily
  Denton, Seyed Kamyar~Seyed Ghasemipour, Burcu~Karagol Ayan, S~Sara Mahdavi,
  Rapha~Gontijo Lopes, et~al.
\newblock Photorealistic text-to-image diffusion models with deep language
  understanding.
\newblock {\em arXiv preprint arXiv:2205.11487}, 2022.

\bibitem{taichi}
Aliaksandr Siarohin, St\'{e}phane Lathuili\`{e}re, Sergey Tulyakov, Elisa
  Ricci, and Nicu Sebe.
\newblock First order motion model for image animation.
\newblock In {\em Advances in Neural Information Processing Systems}, 2019.

\bibitem{monkeynet}
Aliaksandr Siarohin, Stéphane Lathuilière, Sergey Tulyakov, Elisa Ricci, and
  Nicu Sebe.
\newblock Animating arbitrary objects via deep motion transfer.
\newblock In {\em IEEE Conference on Computer Vision and Pattern Recognition},
  2019.

\bibitem{siarohin2021motion}
Aliaksandr Siarohin, Oliver Woodford, Jian Ren, Menglei Chai, and Sergey
  Tulyakov.
\newblock Motion representations for articulated animation.
\newblock In {\em IEEE Conference on Computer Vision and Pattern Recognition},
  2021.

\bibitem{makeavideo}
Uriel Singer, Adam Polyak, Thomas Hayes, Xi Yin, Jie An, Songyang Zhang, Qiyuan
  Hu, Harry Yang, Oron Ashual, Oran Gafni, et~al.
\newblock Make-a-video: Text-to-video generation without text-video data.
\newblock {\em arXiv preprint arXiv:2209.14792}, 2022.

\bibitem{face2face}
Justus Thies, Michael Zollhofer, Marc Stamminger, Christian Theobalt, and
  Matthias Nie{\ss}ner.
\newblock Face2face: Real-time face capture and reenactment of rgb videos.
\newblock In {\em IEEE Conference on Computer Vision and Pattern Recognition},
  2016.

\bibitem{MoCoGANHD}
Yu Tian, Jian Ren, Menglei Chai, Kyle Olszewski, Xi Peng, Dimitris N.~Metaxas,
  and Sergey Tulyakov.
\newblock A good image generator is what you need for high-resolution video
  synthesis.
\newblock In {\em International Conference on Learning Representations}, 2021.

\bibitem{MoCoGAN}
Sergey Tulyakov, Ming-Yu Liu, Xiaodong Yang, and Jan Kautz.
\newblock {MoCoGAN}: Decomposing motion and content for video generation.
\newblock In {\em IEEE Conference on Computer Vision and Pattern Recognition},
  2018.

\bibitem{fvd}
Thomas Unterthiner, Sjoerd van Steenkiste, Karol Kurach, Raphael Marinier,
  Marcin Michalski, and Sylvain Gelly.
\newblock Towards accurate generative models of video: A new metric \&
  challenges.
\newblock {\em arXiv preprint arXiv:1812.01717}, 2018.

\bibitem{hierarchical}
Ruben Villegas, Dumitru Erhan, Honglak Lee, et~al.
\newblock Hierarchical long-term video prediction without supervision.
\newblock In {\em International Conference on Machine Learning}, 2018.

\bibitem{voynov2022sketch}
Andrey Voynov, Kfir Aberman, and Daniel Cohen-Or.
\newblock Sketch-guided text-to-image diffusion models.
\newblock {\em arXiv preprint arXiv:2211.13752}, 2022.

\bibitem{hrnet}
Jingdong Wang, Ke Sun, Tianheng Cheng, Borui Jiang, Chaorui Deng, Yang Zhao,
  Dong Liu, Yadong Mu, Mingkui Tan, Xinggang Wang, et~al.
\newblock Deep high-resolution representation learning for visual recognition.
\newblock {\em IEEE Transactions on Pattern Analysis and Machine Intelligence},
  2020.

\bibitem{ssim}
Zhou Wang, Alan~C Bovik, Hamid~R Sheikh, and Eero~P Simoncelli.
\newblock Image quality assessment: from error visibility to structural
  similarity.
\newblock {\em IEEE Transactions on Image Processing}, 2004.

\bibitem{tuneavideo}
Jay~Zhangjie Wu, Yixiao Ge, Xintao Wang, Weixian Lei, Yuchao Gu, Wynne Hsu,
  Ying Shan, Xiaohu Qie, and Mike~Zheng Shou.
\newblock Tune-a-video: One-shot tuning of image diffusion models for
  text-to-video generation.
\newblock {\em arXiv preprint arXiv:2212.11565}, 2022.

\bibitem{yang2022diffusionsg}
Ling Yang, Zhilin Huang, Yang Song, Shenda Hong, Guohao Li, Wentao Zhang, Bin
  Cui, Bernard Ghanem, and Ming-Hsuan Yang.
\newblock Diffusion-based scene graph to image generation with masked
  contrastive pre-training.
\newblock {\em arXiv preprint arXiv:2211.11138}, 2022.

\bibitem{yang2022diffusion}
Ruihan Yang, Prakhar Srivastava, and Stephan Mandt.
\newblock Diffusion probabilistic modeling for video generation.
\newblock {\em arXiv preprint arXiv:2203.09481}, 2022.

\bibitem{lpips}
Richard Zhang, Phillip Isola, Alexei~A Efros, Eli Shechtman, and Oliver Wang.
\newblock The unreasonable effectiveness of deep features as a perceptual
  metric.
\newblock In {\em IEEE Conference on Computer Vision and Pattern Recognition},
  2018.

\end{thebibliography}
